\documentclass[conference]{IEEEtran}
\IEEEoverridecommandlockouts

\usepackage{cite}
\usepackage{amsmath,amssymb,amsfonts}
\usepackage{algorithmic}
\usepackage{graphicx}
\usepackage{textcomp}
\usepackage{xcolor}
\usepackage{multirow}
\usepackage{enumitem}
\usepackage{color}
\usepackage{booktabs}

\def\BibTeX{{\rm B\kern-.05em{\sc i\kern-.025em b}\kern-.08em
    T\kern-.1667em\lower.7ex\hbox{E}\kern-.125emX}}

\title{Utilizing the Mean Teacher with Supcontrast Loss for Wafer Pattern Recognition
}

\author{
\IEEEauthorblockN{Qiyu Wei\IEEEauthorrefmark{1}, Xun Xu\IEEEauthorrefmark{2}, Zeng Zeng\IEEEauthorrefmark{1}, Xulei Yang\IEEEauthorrefmark{2}}
\IEEEauthorblockA{\IEEEauthorrefmark{1}School of Microelectronics, Shanghai University, Shanghai, China}
\IEEEauthorblockA{\IEEEauthorrefmark{2}Institute for Infocomm Research, A*Star, Singapore, Singapore}
}

\begin{document}



\maketitle


\begin{abstract}
The patterns on wafer maps play a crucial role in helping engineers identify the causes of production issues during semiconductor manufacturing. In order to reduce costs and improve accuracy, automation technology is essential, and recent developments in deep learning have led to impressive results in wafer map pattern recognition. 
In this context, inspired by the effectiveness of semi-supervised learning and contrastive learning methods, we introduce an innovative approach that integrates the Mean Teacher framework with the supervised contrastive learning loss for enhanced wafer map pattern recognition. Our methodology not only addresses the nuances of wafer patterns but also tackles challenges arising from limited labeled data. To further refine the process, we address data imbalance in the wafer dataset by employing SMOTE and under-sampling techniques. 
We conduct a comprehensive analysis of our proposed method and demonstrate its effectiveness through experiments using real-world dataset WM811K obtained from semiconductor manufacturers. Compared to the baseline method, our method has achieved 5.46\%, 6.68\%, 5.42\%, and 4.53\% improvements in Accuracy, Precision, Recall, and F1 score, respectively.
\end{abstract}


\section{Introduction}

Over the past few decades, global demand for semiconductor products has been steadily increasing, and the semiconductor industry has undergone remarkable development following the trajectory set forth by Moore's Law.  As we inch closer to the physical boundaries of device miniaturization, the demands on quality and performance in semiconductor manufacturing processes have intensified unprecedentedly. Enhancing wafer manufacturing yields has emerged as an intricate and challenging endeavor~\cite{theodosiou2023review}. In this context, Artificial intelligence is progressively embraced by the semiconductor industry, with sundry applications spanning various phases of the semiconductor manufacturing process\cite{kim2022advances,choi2023boosted,wei2023wafer}. Some work has been done to suggest practical ways to address this point. For example, Tsai \emph{et al.}~\cite{tsai2020light}proposed a wafer map data augmentation and failure pattern recognition method, Nakazawa \emph{et al.}~\cite{nakazawa2018wafer} proposed a method for wafer map failure pattern recognition using convolutional neural networks, Hao \emph{et al.}~\cite{sheng2024efficient} proposed deep learning framework for mixed-type wafer map defect pattern recognition, and these methods have obtained competitive results.

In real-world fab production scenarios, vast amounts of data are generated daily, and it is impractical to label all of it manually. Most existing studies have only employed the labeled portion of the dataset, leaving a significant amount of unlabeled data unused. Some semi-supervised learning methods, such as Semi-supervised Clustering~\cite{bair2013semi}, Pseudo Label~\cite{lee2013pseudo}, and Temporal Ensembling~\cite{laine2016temporal} have been used in industry. In this work, we utilize the mean teacher~\cite{tarvainen2017mean} network to leverage this large volume of unlabeled data to enhance our model's performance. Compared to other Semi-supervised methods, the Mean Teacher offers a more stable teacher model due to its moving average mechanism and consistency regularization.

This work is also inspired by the recent advancements in the field of contrastive learning~\cite{chen2020simple}, particularly its efficiency in feature extraction. Building on this, we propose the integration of supervised contrastive learning strategies into the mean teacher semi-supervised learning framework. Our concept is that by combining the powerful feature extraction capabilities of supervised contrastive learning with the robustness of the mean teacher model, we can more effectively enhance the model's performance in handling partially labeled data, strengthen the model's ability to capture key features, and improve its adaptability in a semi-supervised learning environment. Through this innovative combination, our approach not only increases learning efficiency but also brings new perspectives and possibilities to the field of semi-supervised learning.

To sum up, the contribution of this work lies in three folds:
\begin{itemize}
\setlength{\itemsep}{0pt}
\setlength{\parsep}{0pt}
\setlength{\parskip}{0pt}
\item We utilize the mean teacher algorithm with a supervised contrast learning method, leveraging a large amount of unlabeled data to enhance the performance of the model.
\item To rectify the issue of data imbalance, we have employed both up-sampling and down-sampling techniques.
\item A comprehensive comparison illustrates the efficacy of our method on the WM811K dataset.
\end{itemize}

\section{Related Works}

\subsection{Recognition Algorithms In Wafer Datasets}
\label{seb:2.1}

Computer vision and machine learning techniques have gained prominence in wafer map recognition, addressing its challenges. Ishida \emph{et al.}~\cite{ishida2019deep} introduced a deep learning framework that autonomously discerns failure pattern characteristics. Yu \emph{et al.}~\cite{yu2015wafer} leveraged manifold learning for wafer map failure detection and recognition. Meanwhile, Shim \emph{et al.}~\cite{shim2020active} introduced an iterative active learning-based CNN for wafer map pattern recognition, enhancing model performance progressively. Jaewoong {et al.}~\cite{shim2023learning} proposed a method of learning from single-defect wafer maps to classify mixed-defect wafer maps. Guangyuan {et al.}~\cite{deng2024efficient} proposed an efficient wafer defect pattern recognition method based on light-weight neural network.
In addition to supervised methods, researchers have begun to explore the application of self-supervised and semi-supervised methods in recognizing failure patterns in wafer graphs. 
Siyamalan ~\cite{manivannan2024semi} proposed Semi-supervised imbalanced classification using a Dual-Head CNN for wafer bin map defects detection.
WaPIRL~\cite{misra2020self} pioneered the application of the self-supervised learning PIRL model in the semiconductor domain. Geng \emph{et al.}~\cite{geng2021wafer} applied few-shot learning to wafer pattern recognition, while Xu \emph{et al.}~\cite{xu2023unsupervised} used unsupervised learning to obtain a feature representation and subsequently constructed a classification task based on those features.
Although deep neural networks (DNNs) have shown promise, challenges like class imbalance and limited labeled data persist. As a remedy, Yu \emph{et al.}~\cite{yu2021multiple} presented a multigranularity generative adversarial network (GAN) for wafer map enhancement. 
However, these methods have drawbacks. The limitations of self-supervised learning, such as with the PIRL model, lie in its utilization of unlabeled data. Its performance may be affected if the self-supervised task doesn't align with the actual task. Iterative active learning might require multiple rounds of user interactions and annotations, which might be impractical in real-world applications.

\subsection{Contrastive Learning}
\label{seb:2.2}

Contrastive learning methods for unsupervised visual representation learning have reached remarkable levels of performance. Chen \emph{et al.}~\cite{chen2020simple} present SimCLR: a simple framework for contrastive learning of visual representations. THEY simplify recently proposed contrastive self-supervised learning algorithms without requiring specialized architectures or a memory bank. Wang \emph{et al.}~\cite{wang2020understanding} identify two key properties related to contrastive loss: (1) alignment (closeness) of features from positive pairs, and (2) uniformity of the induced distribution of the (normalized) features on the hypersphere, directly optimizing for these two metrics leads to better performance. Le-Khac \emph{et al.}~\cite{le2020contrastive} provide a general Contrastive Representation Learning framework.  Robinson \emph{et al.}~\cite{robinson2020contrastive} introduce an unsupervised method for sampling hard negatives for contrastive learning. Based on the remarkable success of these unsupervised contrastive learning methods, some researchers have studied the mechanism of contrastive loss. Wang \emph{et al.}~\cite{wang2021understanding} concentrate on the understanding of the behaviors of unsupervised contrastive loss. 
Contrastive learning, especially self-supervised contrastive learning (SSCL), has achieved great success in extracting powerful features.

\section{Proposed Methodology}
\label{sec:Proposed Methodology}

\begin{figure*}[!htbp] 
\centering 
\includegraphics[width = 0.9\textwidth]{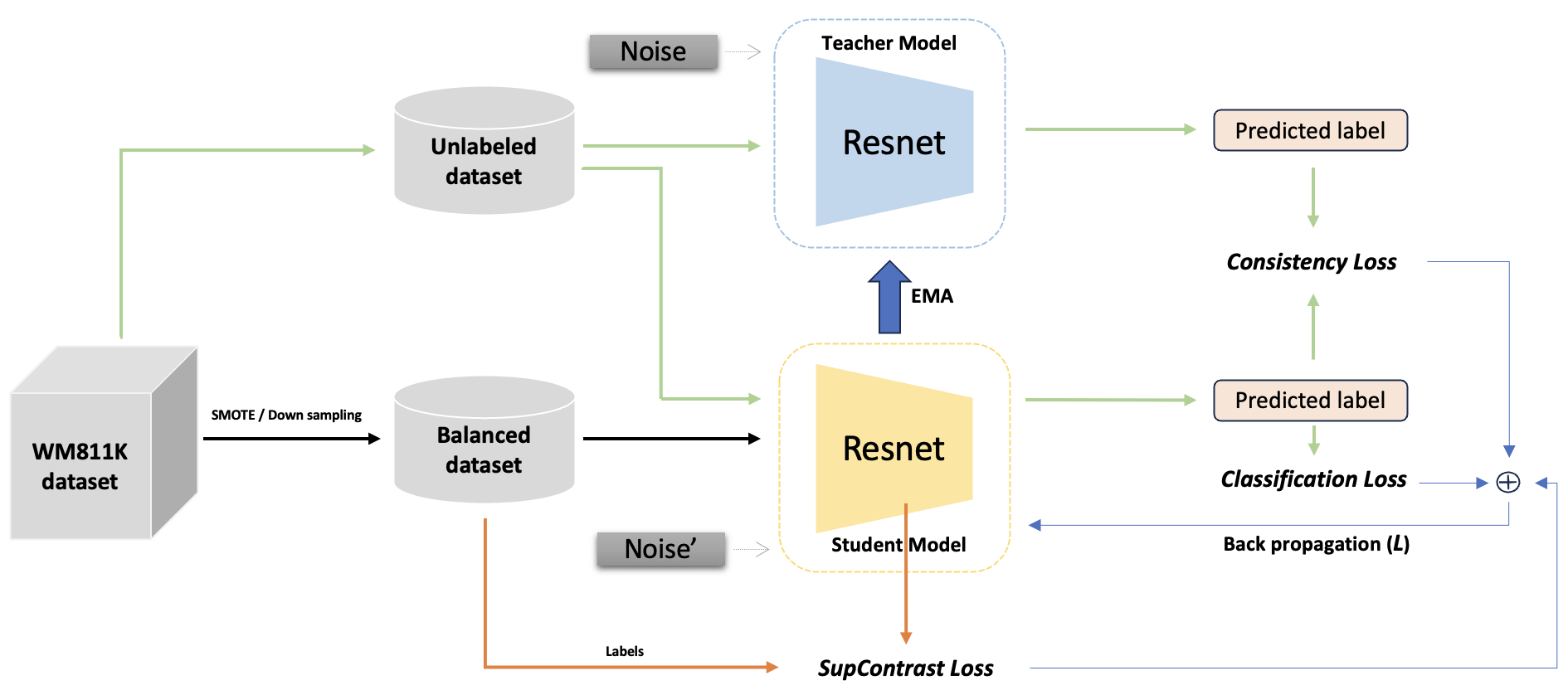}
\caption{
Illustration of Mean Teacher Framework with supercontrast loss. 
}
\label{fig:Mean_heatmap}
\vspace{-0.1in}
\end{figure*}

Wafer map pattern recognition is challenging, and given that many wafer images are unlabeled, semi-supervised learning methods have become a necessary choice. To effectively address this issue, we adopted the Mean Teacher framework with supervised contrastive loss. Fig. \ref{fig:Mean_heatmap} illustrates our proposed architecture for wafer map pattern recognition.

\subsection{Mean Teacher Framework}

Mean Teacher is an advanced semi-supervised learning method that has demonstrated outstanding performance and immense potential in various scenarios, especially in wafer map pattern recognition. 

\subsubsection{Framework Overview}
The underlying principle of the Mean Teacher algorithm is to ensure prediction consistency between two distinct neural network instances: the student and the teacher. While the student model undergoes iterative updates during each training epoch, the teacher model's parameters are governed by an Exponential Moving Average (EMA) of the student model. This design ensures the teacher model's relative stability across extensive training epochs. When the same unlabeled data is fed into both networks, we want the outputs of both to be as identical as possible so that a large amount of unlabeled data can be utilized to help with model training. The ability to utilize large amounts of unlabeled data is one of the main advantages of Average Teacher.


\subsubsection{Training Strategies}
In the average teacher model, the student model and the teacher model are trained using two different strategies. In this case, labeled data is used to train only the student model. Subsequently, at each step, a very small amount of weights from the student model is assigned to the teacher model, called exponential moving average weights. Introducing noise during training allows the model to learn more universal features, so that the model can still perform well in real applications when faced with data that is different from the training set.

\textit{Student Model Update}: A classification loss can be obtained by taking the wafer image as input and getting the prediction through the student model. Then using unlabeled data, a consistency loss measure is computed by comparing the outputs of the student model and the teacher model, and this loss indicates the difference in prediction between the two models. The two losses were summed and back-propagated to update the parameters of the student model.

\textit{Teacher Model Update and EMA}: Concurrently with the student's updates, the teacher model's parameters undergo evolutionary adjustments governed by the Exponential Moving Average (EMA). Mathematically, the EMA for the teacher model's parameters at time \( t \), denoted \( \theta_{\text{teacher}}(t) \), is defined as:
\begin{equation}
\theta_{\text{teacher}}(t) = \alpha \cdot \theta_{\text{teacher}}(t-1) + (1 - \alpha) \cdot \theta_{\text{student}}(t)
\end{equation}
Where \( \alpha \) is a decay factor within [0,1]. The adoption of EMA ensures a smoother learning trajectory for the teacher model, with \( \alpha \) determining the weightage of past observations. This EMA-driven approach bestows the teacher model with enhanced stability and the capability to leverage historical learning.

\subsection{Supervised Contrastive Learning}
\label{seb:3.2}

In our research, we applied the concept of Supervised Contrastive Learning (SCL) loss, as outlined in \cite{khosla2020supervised}. We hypothesize that contrastive learning provides superior guidance for intermediate layers compared to traditional task-specific loss supervision. Typically, this method identifies two augmentations from the same image as a 'positive pair,' and those from different images as 'negative pairs'. The training process involves teaching the neural network to reduce the distance between positive pairs while increasing it for negative pairs. This enables the network to adapt to a range of data enhancements, such as color variations and random grayscale transformations. Given that these enhancements are often low-level, task-agnostic, and adaptable to a wide range of visual tasks, we believe they offer more valuable insights for the learning of intermediate layers.

This loss function aims to improve the discriminative capabilities of learned representations by utilizing label information. Essentially, it leverages the data's label information to steer the contrastive learning process. The core of Supervised Contrastive Learning is its loss function. Given a batch of data, which includes positive samples (similar samples) and negative samples (dissimilar samples), the loss function can be defined as:

\begin{equation}
L_{supcontrast} = \sum_{i=1}^{N} \frac{-1}{|P(i)|} \sum_{p \in P(i)} \log \frac{\exp(\hat{\mathbf{f}}_i \cdot \hat{\mathbf{f}}_p / \tau)}{\sum_{a \in A(i)} \exp(\hat{\mathbf{f}}_i \cdot \hat{\mathbf{f}}_a / \tau)}
\end{equation}

This formulation uses a temperature-scaled cosine similarity metric, promoting an environment where similar samples are more closely aligned in the feature space, where:
\begin{itemize}
    \item $P(i)$ denotes the set of indices of all positive samples in the batch for the $i$-th sample.
    \item $A(i)$ represents the set of indices for all samples in the batch, including $i$ itself.
    \item $\tau$ is the temperature parameter.
    \item $\hat{\mathbf{f}}_i \cdot \hat{\mathbf{f}}_p$ is the dot product of the normalized feature vectors, serving as a similarity measure. \(\hat{\mathbf{f}}_i\) and \(\hat{\mathbf{f}}_p\) are the normalized feature vectors of samples \(i\) and \(p\), respectively.
\end{itemize}

In the process, we typically adjust the loss by scaling it with $|P(i)|$. Unlike the self-supervised contrastive loss, where a query and a key are considered as a positive pair if they are augmented versions of the same image, the supervised contrastive loss views them as a positive pair if they come from the same category.

Feature representations are obtained through a deep learning model, transforming the input data into a form that is suitable for effective comparison. The goal is to adjust the model parameters so that the feature representations of samples from the same class are brought closer together, while those from different classes are pushed further apart. In supervised settings, a binary mask $\mathbf{M}$ is applied where $M_{ij} = 1$ if samples $i$ and $j$ share the same label, and $0$ otherwise. This mask is used to identify positive pairs in the loss computation. For unsupervised settings, an identity matrix is used as the mask, aligning the loss with the SimCLR formulation.

We add this supervised contrastive loss as additional information to the original loss. Mathematically, the total loss can be represented as:
\begin{flalign}
L = L_{consistency} + L_{classification} + L_{supcontrast}
\end{flalign}
This approach in Supervised Contrastive Learning enables the model not just to differentiate between different categories of samples, but also to understand the nuances within the same category.

\section{Experiments}

\subsection{Experimental Settings}

We conducted experiments on the WM-811K wafer map dataset \cite{wu2014wafer}, which is a public dataset widely used in semiconductor manufacturing research. It includes 811,457 wafer map images from 46,294 lots, with 172,950 manually labeled, and contains nine patterns. The original dataset has a significant imbalance. The Non-Pattern pattern accounts for a vast majority, while the Donut and NearFull patterns only make up $0.3\%$ and $0.1\%$, respectively. To address this issue, we have explored both over-sampling and under-sampling techniques. We use the Synthetic Minority Over-sampling Technique (SMOTE) \cite{chawla2002smote} on the training set to generate synthetic samples by interpolating between neighboring samples of the minority, then the distribution of the data set is well-balanced.

For implementation, we utilized a mean teacher algorithm framework with a ResNet18-based classification network. The training was conducted on 10\% of the WM811K dataset, with the remaining data serving as unlabeled input for both student and teacher networks, ensuring a fair comparison.
In this paper, we employ four widely used evaluation metrics, namely Precision, Recall, F1 score, and Accuracy, to assess the results of our experiments.

\subsection{Performance study}

\begin{table}[!t]
\caption{Overall Experiment Results}
\centering
\label{tab:overall}
\begin{tabular}{l c}
\toprule
Methods                             & Overall Results \\ &( Accuracy, Precision, Recall, F1 ) \\ \hline
Resnet (Baseline)           & 79.17\%, 79.56\%, 78.99\%, 78.87\%                  \\ 
{+ mean teacher }              & 81.14\%, 82.55\%, 81.15\%, 81.29\%               \\
{+ SupConLoss  }               & 84.13\%, 85.53\%, 83.37\%, 82.98\%               \\
{+ mean teacher \& SupConLoss}  & \textbf{84.63\%, 86.24\%, 84.41\%, 83.40\%  }   \\
\toprule
\end{tabular}
\end{table}

\begin{table}[!t]
\centering
\caption{Experiment Results Details}
\label{tab:result}
\begin{tabular}{p{1.75cm}|p{1.25cm}|c}
\toprule
\textbf{Method} & \textbf{Class} & \textbf{Accuracy, Precision, Recall, F1} \\
\midrule
\multirow{10}{*}{Resnet(Baseline)} 
& Center & 97.67\%, 88.97\%, 89.38\%, 89.17\% \\
& Donut & 97.72\%, 92.55\%, 89.53\%, 91.02\% \\
& Edge-Loc & 89.03\%, 71.75\%, 51.02\%, 59.63\% \\
& Edge-Ring & 98.38\%, 95.80\%, 92.78\%, 94.27\% \\
& Loc & 89.38\%, 52.56\%, 48.15\%, 50.26\% \\
& Near-full & 100.00\%, 92.36\%, 100.00\%, 96.03\% \\
& Random & 96.76\%, 94.38\%, 86.30\%, 90.16\% \\
& Scratch & 94.29\%, 57.14\%, 74.61\%, 64.72\% \\
& None & 95.09\%, 70.55\%, 79.18\%, 74.62\% \\
\midrule
\multirow{10}{*}{+ Mean Teacher} 
& Center & 96.11\%, 95.48\%, 83.15\%, 88.89\% \\
& Donut & 98.53\%, 87.50\%, 93.11\%, 90.22\% \\
& Edge-Loc & 91.20\%, 78.72\%, 59.82\%, 67.98\% \\
& Edge-Ring & 99.19\%, 93.56\%, 96.22\%, 94.87\% \\
& Loc & 90.80\%, 67.24\%, 56.46\%, 61.38\% \\
& Near-full & 100.00\%, 98.85\%, 100.00\%, 99.42\% \\
& Random & 96.56\%, 96.33\%, 85.28\%, 90.47\% \\
& Scratch & 96.01\%, 56.84\%, 82.41\%, 67.27\% \\
& None & 93.88\%, 68.46\%, 73.92\%, 71.09\% \\
\midrule
\multirow{10}{*}{+ SupConLoss} 
& Center & 97.98\%, 67.80\%, 90.91\%, 77.67\% \\
& Donut & 95.47\%, 76.67\%, 71.88\%, 74.19\% \\
& Edge-Loc & 90.93\%, 85.71\%, 62.50\%, 72.29\% \\
& Edge-Ring & 98.99\%, 100.00\%, 94.74\%, 97.30\% \\
& Loc & 98.49\%, 69.23\%, 92.31\%, 79.12\% \\
& Near-full & 100.00\%, 97.78\%, 100.00\%, 98.88\% \\
& Random & 98.49\%, 82.81\%, 94.64\%, 88.33\% \\
& Scratch & 98.49\%, 94.55\%, 94.55\%, 94.55\% \\
& None & 89.42\%, 95.24\%, 48.78\%, 64.52\% \\
\midrule
\multirow{10}{*}{\shortstack{+ Mean Teacher \\ \& SupConLoss}} 
& Center & 95.97\%, 67.92\%, 81.82\%, 74.23\% \\
& Donut & 98.99\%, 91.84\%, 95.74\%, 93.75\% \\
& Edge-Loc & 86.40\%, 100.00\%, 38.64\%, 55.74\% \\
& Edge-Ring & 99.50\%, 81.36\%, 97.96\%, 88.89\% \\
& Loc & 98.49\%, 82.93\%, 91.89\%, 87.18\% \\
& Near-full & 100.00\%, 88.10\%, 100.00\%, 93.67\% \\
& Random & 99.50\%, 84.21\%, 97.96\%, 90.57\% \\
& Scratch & 94.46\%, 90.00\%, 71.05\%, 79.41\% \\
& None & 95.97\%, 89.80\%, 84.62\%, 87.13\% \\
\bottomrule
\end{tabular}
\end{table}

In our comparative study, we established ResNet18 as the baseline and conducted multiple control experiments, including combinations of ResNet with SupConLoss, mean teacher, and mean teacher with SupConLoss. As illustrated in Table \ref{tab:overall}, The experimental results demonstrated that the combination of Mean Teacher and SupConLoss yields the best performance. This finding proves that combining Mean Teacher with SupConLoss enhances learning efficiency, this combination achieved improvements of 5.46\%, 6.68\%, 5.42\%, and 4.53\% across the four metrics, respectively.

The ablation study further clarified the individual contributions of SupConLoss and the Mean Teacher method. A comparative analysis between Mean Teacher + SupConLoss and Mean Teacher only revealed an overall enhancement of 2.11\% in the F1 score. As detailed in Table \ref{tab:result} , the inclusion of SupConLoss consistently improved performance, particularly in categories like "Donut", "Loc", "Scratch", and "None", suggesting that the combined model might be more effective in handling complex features, SupConLoss potentially provides a more robust feature space for classification tasks, thereby adding significant value to the learning process.

When combined with SupConLoss, the ResNet model showed a 4.11\% improvement in the F1 score across most categories compared to ResNet alone. This was particularly evident in categories like "Edge-Ring" and "Near-full", signifying that SupConLoss can significantly enhance performance in specific scenarios. This suggests that SupConLoss has a certain level of applicability and effectiveness across different model structures.

Regarding the Mean Teacher method, it achieved a 2.42\% improvement in the F1 score compared to the standalone ResNet. The detailed data presented in Table \ref{tab:result} reveals significant performance enhancements of the mean teacher model over the solo ResNet in most categories. For instance, in categories like "Edge-Loc", "Edge-Ring", "Loc" and "Near-full". The mean teacher can enhance the model's generalization capabilities on unlabeled data through smoothed labels and model outputs, thereby bolstering overall model performance. 

\section{Conclusion}

In this work, we presented a simple but efficient approach for wafer defect classification, capitalizing on the strengths of the Mean Teacher framework and supervised contrastive loss. Our methodology effectively addresses the intricacies of wafer patterns and the challenges stemming from limited labeled data.  Quantitative evaluations corroborated the superior performance of our approach over traditional methods.

\clearpage

\bibliographystyle{IEEEtran}
\bibliography{mybib}

\end{document}